\documentclass[letterpaper]{article} 
\usepackage{aaai25}
\usepackage{times}  
\usepackage{helvet}  
\usepackage{courier}  
\usepackage[hyphens]{url}  
\usepackage{graphicx} 
\usepackage{natbib}  
\usepackage{caption} 

\usepackage{microtype}
\usepackage{algorithm}
\usepackage{algorithmic}

\usepackage{booktabs}
\usepackage{caption}
\usepackage{amsthm}
\usepackage{algorithm}
\usepackage{algorithmic}
\usepackage{amsfonts}
\usepackage{subcaption}
\usepackage{booktabs} 
\usepackage{amsmath}
\usepackage{enumitem}
\usepackage{comment}
\usepackage{graphicx}
\usepackage{bm}
\usepackage{mathrsfs}
\usepackage{threeparttable}
\usepackage{multirow}
\usepackage{mathtools}
\usepackage{xurl} 

\usepackage{newfloat}
\usepackage{listings}
\DeclareCaptionStyle{ruled}{labelfont=normalfont,labelsep=colon,strut=off} 
\floatstyle{ruled}
\newfloat{listing}{tb}{lst}{}
\floatname{listing}{Listing}
\usepackage{xcolor}
\definecolor{lightblue}{RGB}{0,120,215}

\title{Certified Causal Defense with Generalizable Robustness}

\author {
    Yiran Qiao\textsuperscript{\rm 1},
    Yu Yin\textsuperscript{\rm 1},
    Chen Chen\textsuperscript{\rm 2}
    Jing Ma\textsuperscript{\rm 1}
}
\affiliations {
    \textsuperscript{\rm 1}Case Wester Reserve University\\
    \textsuperscript{\rm 2}University of Virginia\\
    \{yxq350, yxy1421, jing.ma5\}@case.edu, zrh6du@virginia.edu
}

\usepackage{bibentry}

\begin{document}

\maketitle
\newcommand{\mymodel}{GLEAN}

\begin{abstract}
\looseness=-1
While machine learning models have proven effective across various scenarios, it is widely acknowledged that many models are vulnerable to adversarial attacks. Recently, there have emerged numerous efforts in adversarial defense. Among them, certified defense is well known for its theoretical guarantees against arbitrary adversarial perturbations on input within a certain range (e.g.,  $l_2$ ball). However, most existing works in this line struggle to generalize their certified robustness in other data domains with distribution shifts. This issue is rooted in the difficulty of eliminating the negative impact of spurious correlations on robustness in different domains. To address this problem, in this work, we propose a novel certified defense framework GLEAN, 
which incorporates a causal perspective into the generalization problem in certified defense. More specifically, our framework integrates a certifiable causal factor learning component to disentangle the causal relations and spurious correlations between input and label, and thereby exclude the negative effect of spurious correlations on defense. On top of that, we design a causally certified defense strategy to handle adversarial attacks on latent causal factors. In this way, our framework is not only robust against malicious noises on data in the training distribution but also can generalize its robustness across domains with distribution shifts. Extensive experiments on benchmark datasets validate the superiority of our framework in certified robustness generalization in different data domains. Code is available at \textcolor{lightblue}{\textit{https://github.com/yrqiao/Glean/tree/main}}.

\end{abstract}


\section{Introduction}

Machine learning (ML) models, particularly deep neural networks (DNN), have demonstrated remarkable success across many areas \cite{devlin2018bert,silver2017mastering,he2016deep}. Despite their success, these models still exhibit significant vulnerabilities to adversarial perturbations on input \cite{szegedy2013intriguing,goodfellow2014explaining,biggio2013evasion}. A typical example in image classification is that a trained classifier that correctly classifies an image $x$ can be easily fooled by a perturbed image $x+\delta$, where $\delta$ represents adversarial perturbations imperceptible to human perceptions. This weakness impedes the deployment of ML models in critical applications where security and reliability are priorities, such as autonomous driving and healthcare. 

\looseness=-1
In the past few decades, researchers have developed numerous defense methods to enhance the adversarial robustness of ML models. Many of them are based on \textit{adversarial training} \cite{goodfellow2014explaining,madry2017towards,zhang2019theoretically,athalye2018obfuscated}, which incorporates adversarial samples into model training. Despite its impressive performance, adversarial training is an empirical approach that lacks theoretical guarantees. That is, although it can enhance robustness against certain types of attacks, it may still be vulnerable to other unknown, or more potent adversarial perturbations. Differently, another line of work develops \textit{certified robustness}. A certified robust classifier can theoretically guarantee that its prediction for a point $x$ remains constant within a certain specified range (a.k.a. \textit{radius}) of perturbations on $x$, regardless of the type of attack. Randomized smoothing-based certified defense \cite{lecuyer2019certified,li2018second,cohen2019certified} is one of the most representative methods in this area. Specifically, given an arbitrary base classifier $f$, this method can convert it to a certifiably robust classifier $g$, which is created by randomly sampling multiple noised versions of a given input and using the aggregated output from these variations to make final predictions.
Inspired by this approach, many subsequent studies \cite{li2019certified,jeong2020consistency,jeong2021smoothmix,salman2019provably,zhai2020macer} have expanded upon the basis of random smoothing. 

Currently, most existing certified defense 
works focus on data in the same domain yet overlook other domains with distribution shifts. 
This limitation can result in a markedly degraded certified robustness performance when these methods are applied to the test domain \cite{sun2021certified}. As discussed in previous work \cite{ilyas2019adversarial,beery2018recognition}, such degradation of robustness lies in the fact that ML models tend to overfit spurious correlations between features and labels. As these spurious correlations often vary across different domains \cite{ye2022ood},
fitting spurious correlations can easily lead $g$ to make incorrect predictions or correct predictions but with lower confidence levels. The former results in the certified radius being assigned zero, while the latter also leads to a reduced certified radius. Therefore, domain shifts can lead to weak generalization w.r.t. not only prediction performance but also certified robustness.
Different from ML models, humans can naturally capture the invariant relations between labels and their corresponding causal factors, various studies \cite{zhang2020causal,tang2020long,scholkopf2021toward} argue that human's inherent causal view brings a solution to avoid the domain generalization hurdle for robustness. 
Inspired by this, in this paper, we study the problem of \textit{generalizing the certified robustness under domain shifts from a causal view}.



\looseness=-1
However, addressing this problem presents multifaceted challenges. \textbf{Challenge 1:} 
As aforementioned, spurious correlations varying across domains adversely affect robustness. To achieve robustness across domains, it is crucial to effectively remove the impact of these spurious correlations. However, identifying and eliminating the impact of spurious correlations on robustness in different domains presents a challenge. \textbf{Challenge 2:} Apart from spurious factors, the distribution shifts often bring challenges for the model to defend against the perturbations on the factors that causally determine the label in unseen domains, which leads to diminished certified robustness. 
\textbf{Challenge 3:} It is important to provide theoretical guarantees for robustness on other data domains, but most existing works remain in empirical observations and lack theoretical analysis. Although some work in certified defenses has provided upper bounds on perturbations while maintaining robustness, they were not designed to address certified robustness in the domain shift context.

\looseness=-1
In this work, to tackle these challenges, we propose a novel framework \underline{\textbf{G}}enera\underline{\textbf{L}}izable c\underline{\textbf{E}}rtified c\underline{\textbf{A}}usal defe\underline{\textbf{N}}se (\textbf{\mymodel}) that enhances the certified robustness of models on data in different domains. To mitigate the influence of spurious correlations on robustness generalization (Challenge 1), we construct a causal model for data in different domains, and simultaneously conduct a causal analysis on model robustness and generalization. Based on the causal model, we filter out the impact from spurious correlations and enhance robustness across domains. This is different from most existing defense algorithms which take the same strategy indiscriminately towards all the input features. 
To achieve certified robustness through causal factors (Challenges 2), we utilize a certified causal factor learning module with Lipschitz constraint. 
This module enables certification through the latent representation space for high-level causal factors, conducting certified defense for perturbations on causal factors that determine the label in different domains. 
To bring theoretical guarantees for robustness on different data domains (Challenge 3), we derive a theoretical analysis by leveraging the theoretical support of certified defense and causal inference. Our main contributions can be summarized as:
\begin{itemize}[topsep=0pt]
\item We investigate an important but underexplored problem of certified defense on data in different domains. We analyze the significance of this research problem and its corresponding challenges.
\item We propose a novel causality-inspired framework \mymodel~for this problem, extending certified robustness across data domains. Specifically, we develop a certified defense strategy based on certifiable causal factor learning, which excludes spurious correlations and provides a certified radius for test data with a theoretical guarantee. 
\item We conduct extensive experiments to evaluate our framework on both synthetic and real-world datasets. The results show that our framework significantly outperforms the prevalent baseline methods.
\end{itemize}

\section{Preliminaries and Related Work}

We consider a classification task with $x$ representing an input instance and $y$ denoting the corresponding label, where $x\in\mathcal{X}, y\in\mathcal{Y}:=\{1,...,K\}$, $\mathcal{X}$ and $\mathcal{Y}$ represent the input space and the label space, respectively. A classifier trained for this task can be denoted by $f:\mathcal{X}\rightarrow \mathcal{Y}$. The data may be collected from different domains (i.e., environments). We use the superscript $(\cdot)^d$ to denote the data in a certain domain $d$. 

\subsection{Certified Defense}
\looseness=-1
\subsubsection{Robust Radius} 
\looseness=-1
The robust radius for an instance $x$ is the largest range (e.g., a $l_2$ ball) centered at $x$, within which $f$ provides a correct prediction for $x$, and this prediction remains constant. It is defined as follows:
\begin{equation}
    R(f; x, y) = \min\nolimits_{f(x') \neq f(x)} \|x' - x\|_2.
\end{equation}
Unfortunately, calculating the robust radius for neural networks is proven to be an NP-complete problem \cite{katz2017reluplex,sinha2017certifying} thus both challenging and time-consuming. 
\subsubsection{Certified Radius}
Many previous works proposed certification methods to derive a \textit{certified radius} that is the lower bound of the robust radius. Researches in this area fall into two categories: exact methods and conservative methods. Exact methods \cite{ehlers2017formal,bunel2018unified,tjeng2017evaluating}, usually based on Satisfiability Modulo Theories or mixed integer linear programming, guarantee the identification of a perturbation $\delta$ within a radius $r$ that can cause $f$ to change its prediction. However, they require the model to have a limited scale. 
Conservative methods \cite{wong2018provable,wong2018scaling,gowal2018effectiveness} ensure the detection of existing adversarial examples and, in addition, refuse to make certification for some vulnerable data points. These methods, though more scalable, impose specific assumptions on the model's architecture. 

\subsubsection{Randomized Smoothing}  
Randomized Smoothing (RS) is proposed to tackle the above limitations, which can be applied to any architectures~\cite{cohen2019certified}. It constructs a smoothed classifier $g$ from an arbitrary base classifier $f$. The definition of $g$ is as follows:
\begin{equation}
    g(x)=\arg\max\nolimits_{y\in\mathcal{Y}} P(f(x+\eta)=y)
\end{equation}
In this formula, $\eta\sim\mathcal{N}(0,\sigma^2\boldsymbol{I})$ is the isotropic Gaussian noise with the noise level $\sigma$ as the hyperparameter. The smoothed classifier $g$ can be summarized as returning the class most likely to be predicted by $f$ when the input $x$ is sampled from Gaussian distributions. \cite{cohen2019certified} provide the theoretical form of certified radius which is the lower bound of the robust radius:
\begin{equation}
\label{eq3}
    CR(f; x, y)=\frac{\sigma}{2}(\Phi^{-1}(p_A)-\Phi^{-1}(p_B)),
\end{equation}
where $p_A=P(f(x+\eta)=y_A)$, $p_B=\max_{y\neq y_A}P(f(x+\eta)=y)$, meaning that $f$ will mostly return class $y_A$ with the probability $p_A$, and will return the runner-up class with the probability $p_B$. $\Phi^{-1}$ is the inverse of the standard Gaussian cumulative distribution function. Then $g(x+\delta)=y_A$ for all $||\delta||_{2}\leq{CR}$.


\begin{figure}[t]
\centering
\includegraphics[width=0.9\columnwidth,height=2.4in]{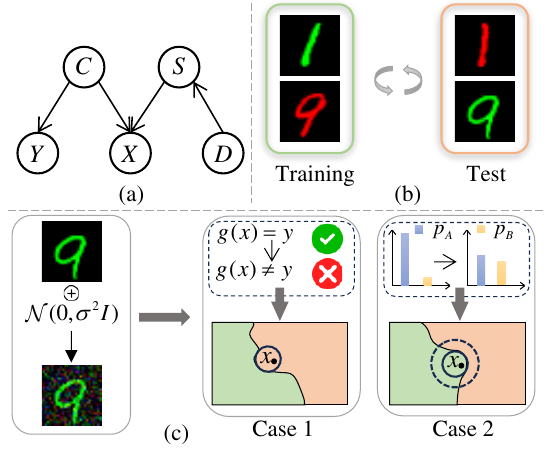}
\vspace{-2mm}
\caption{(a) Causal graph of data generation across domains; (b) A showcase of domain shift, here we use images in CMNIST as an example. (c) Two common cases of domain shifts leading to decreased ACR in certification. The pink area represents an incorrect decision area, green signifies the correct decision area, and circles represent a robust $l_2$ ball.}
\label{fig1}
\vspace{-3mm}
\end{figure}

\subsection{Reduced Certified Robustness in Unseen Domains} There widely exist distribution shifts between data in different domains, i.e., $P^d(X,Y)\ne P^{d'}(X,Y)$, where $d\ne d'$.
Inspired by \cite{zhang2013domain, pearl2016causal}, we use the causal graph shown in Figure \ref{fig1} (a) to illustrate the causal model for data across domains. 
Specifically, as shown in Figure \ref{fig1}, we discuss the causal relations among five variables: label $Y$ (e.g., an object type), input features $X$ (e.g., an image), causal factors $C$ (e.g., the object shape in the image) that determine the label, non-causal factors $S$ (e.g., the background in the image), and domain variable $D$ (e.g., the data source). $C$ and $S$ are usually high-level latent concepts without observed supervision. Noticeably, $S$ often has spurious correlations with $Y$, even if they are not causally related. Such spurious correlations often vary in different domains, i.e., $P^d(Y|S)\ne P^{d'}(Y|S)$.

\looseness=-1
Distribution shift often brings challenges in certified robustness \cite{sun2022spectral}. Here, we use a simple experiment to show the rapid deterioration of certified robustness on data in different domains, where the task is to classify the Colored-MNIST (CMNIST) dataset \cite{arjovsky2019invariant}. CMNIST is a modified version of the handwritten digit image dataset MNIST \cite{lecun1998gradient}, artificially constructed to include two colors, red and green. The colors are strongly but spuriously correlated with the label $Y$ ($Y=0$ for digits $0\sim4$, and $Y=1$ for digits $5\sim 9$).
For this dataset, the color of the digits is a non-causal factor $S$, while the shape of the digits is the causal factor $C$. The spurious correlation between $Y$ and $S$ in the training domain is reversed in the test domain, as shown in Figure \ref{fig1} (b). 

In the CMNIST dataset, it is unsurprising that a classifier relying on the digit color would fail on the test domain due to the shift in the spurious correlation $P(Y|S)$. To show the negative impact of spurious correlation on certified robustness, we compare the Average Certified Radius (ACR, a metric used to evaulate certified robustness) of random smoothing-based certified defense on CMNIST with the results on MNIST (where there is no digit color and thus the above spurious correlations do not exist). As observed from the results in Table \ref{table 1}, there is a significant degradation of prediction accuracy and ACR on the test domain, indicating severe issues for certified defense under domain shift.


\begin{table}[ht]
\centering
\caption{Comparison of the certified defense performance with/without domain shift. The metrics include the prediction accuracy and the Average Certified Radius (ACR).}
\begin{tabular}{ccc}
\toprule
Dataset & Test Acc & ACR ($\sigma=0.25$) \\
\midrule
CMNIST & 21.01\% & 0.07 \\
MNIST & 72.03\% & 0.37 \\
\bottomrule
\end{tabular}
\label{table 1}
\end{table}


\subsection{L-Lipschitz Networks} 
\textbf{Definition 1 (Lipschitz Continuity).} 
\begin{itshape}
A function $f: X\rightarrow{Y}$ is called Lipschitz continuous if there exists a non-negative constant $L$ (known as the Lipschitz constant) such that for all $x_1, x_2\in{X}$ the following condition is met:
\begin{equation}
    ||f(x_{1})-f(x_{2})||_{2}\leq{L}||x_{1}-x_{2}||_{2}.
\end{equation}
\end{itshape}

\looseness=-1
\noindent Based on the definition, if a nerual network $f$ is 1-Lipschitz, then for any input $x$, the output $y$ satisfies $||\mathbf{y}||_2\leq||\mathbf{x}||_2$. Equivalently, $||\mathbf{y_1}-\mathbf{y_2}||_2\leq||\mathbf{x_1}-\mathbf{x_2}||_2$.


\section{\mymodel: Framework and Theories}

\looseness=-1
In this section, we introduce the detailed technologies and theories in our proposed framework. We begin by proposing a causal view of robustness under domain shifts. Next, we introduce our design of a certifiable causal factor learning module to exclude the impact of spurious correlations on robustness. Then, we explain the whole certified defense process through the latent causal space, providing a theoretical guarantee for the certified robustness of data in different domains.

\subsection{Causal View of Robustness and Cross-Domain Generalization}
As introduced in the last section, we illustrate our causal graph in Figure \ref{fig1} (a). Noticeably, although the spurious correlations vary in different domains, since $C\rightarrow{Y}$ has a direct causal link, the relationship between $C$ and $Y$ remains invariant across domains and is thus unaffected by domains. This inspired invariant learning based on the following causal invariance assumption \cite{li2022out}:

\noindent\textbf{Assumption 1 (Causal Invariance over Domain Shifts)}
\begin{itshape}
For any two domains $d$ and $d'$,  the probability $P(Y|C)$ is invariant to domain shifts, i.e.,:
\begin{equation}
    P^{d}(Y|C) = P^{d^{\prime}}(Y|C),\forall d,d^{\prime} \in \mathbb{D},
\end{equation}
where $\mathbb{D}$ is the set of all possible domains.
\end{itshape}

Based on this assumption, a model that can identify causal factors and make predictions based on them can be generalized to unseen domains. 

From a robustness perspective, distribution shifts introduce significant additional challenges. At a high level, robustness can be viewed as a generalization problem over an adversarial distribution \cite{xin2023connection}. This adversarial distribution often differs from the unseen domains derived from natural distributions, necessitating more sophisticated methods to capture high-level causal factors in decision-making while filtering out the impact of adversarial perturbations. More specifically, for a target domain $d'$, we have:
\begin{equation}
    P^{d'}\!(Y|X)\!\!=\!\!\int_{c\in \mathbb{C}}\!\!\! P^{d'}(c|X)P^{d'}(Y|c) \!=\! \!\int_{c\in \mathbb{C}}\!\!\! P^{d'}(c|X)P^{d}(Y|c),
\label{eq:decompose_cause}
\end{equation}
\begin{equation}
    P^{d'}(Y|X)=\int_{s\in \mathbb{S}} P^{d'}(s|X)P^{d'}(Y|X,s),
\end{equation}
where $\mathbb{C}$ and $\mathbb{S}$ are the space of $C$ and $S$, respectively. Here, $d\ne d'$. Each of the equations above decomposes $P(Y|X)$ into two components. As shown in Eq. \ref{eq:decompose_cause}, the model can generalize to a different (or even adversarial) domain $d'$ if it accurately captures the causal factors $C$ from $X$. The other term $P(Y|C)$ remains invariant across domains, which helps to mitigate the risk of increased vulnerability in new domains. However, as indicated by Eq. (7), $P^{d'}(Y|X,s)$ varies across domains, which can increase the vulnerability to adversarial perturbations. This increased vulnerability stems from two main issues, as illustrated in Figure \ref{fig1} (c): 1) reduced accuracy in the test domain, leading to diminished prediction reliability even under slight perturbations; and 2) the change in $P^{d'}(Y|X,s)$ across domains increases decision uncertainty at each $X=x$ due to potential conflicts between $P(Y|C)$ and $P(Y|S)$. These factors collectively complicate the task of achieving robustness across different domains. The above analysis indicates the importance of incorporating a causal view into the robustness problem across domains. In our framework, we identify the causal factors from input (i.e., modeling $P(C|X)$) with certifable robustness, and conduct certified defense based on an invariant predictor $P(Y|C)$.

\begin{figure*}
\centering
  \includegraphics[width=0.9\textwidth]{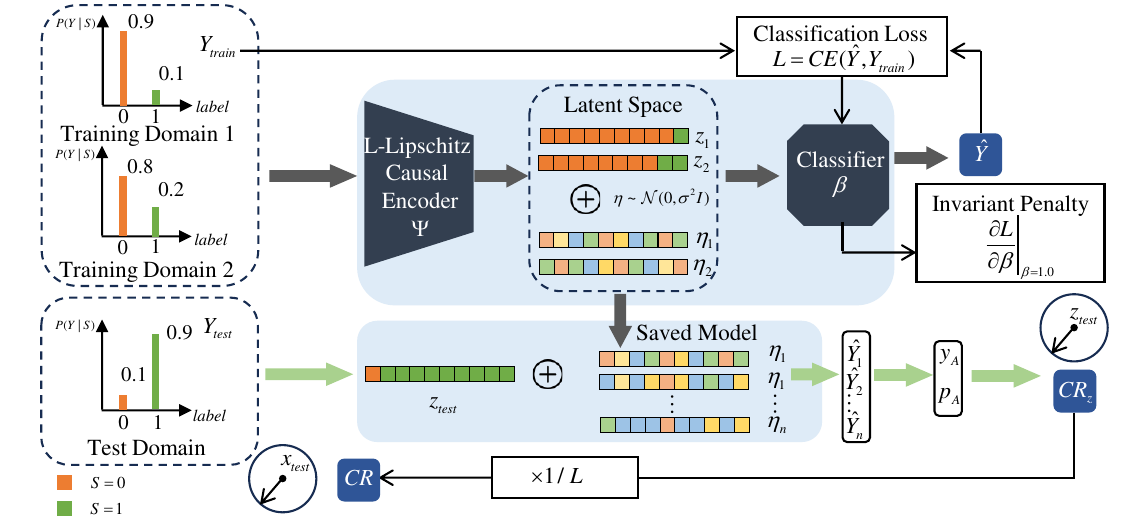}
  \vspace{-2mm}
  \caption{An overview of the proposed framework \mymodel. The upper part represents the training process, while the lower part depicts the certification process on test domain. Here, we showcase two training domains and one test domain with two classes 0 and 1, where the color of the object is a spurious factor. We define $S=0$ as orange and $S=1$ as green. In training domain 1, the spurious distribution between color and category is $P(Y=0|S=0)=0.9$ and $P(Y=1|S=1)=0.1$. These values change to 0.8 and 0.2, respectively in training domain 2, and then to 0.1 and 0.9 in the test domain. Thus, there is a correlation shift between the different domains of this dataset. The causal encoder is equipped with Lipschitz constraints with Lipschitz constant $L$. $Y_{train}$ and $Y_{test}$ are ground truth labels. $\hat{Y}$ is the predicted label. $z$ is the causal latent representations and each $\eta$ is a Gaussian noise. $y_A$ is the most probable class among all the $\hat{Y}$ after sampling with the probability $p_A$. Then we can leverage $p_A$ to compute the certified radius in latent space $CR_z$, and finally revert it back to get the certified radius $CR$ in input space.}
  \label{fig2}
\vspace{-3mm}
\end{figure*}

\subsection{Causal Encoder with Lipschitz Constraint}
Inspired by the above analysis and the observations of the aforementioned toy experiment, to achieve robustness in different domains, we develop a method to robustly identify causal factors from the input for downstream prediction. It is worth mentioning that, for many real-world scenarios, identifying causal factors in the input space (e.g., image pixels) is difficult without segmentation labels, and also less meaningful, because causal factors are often high-level concepts. Therefore, our method is built upon a representation space, where we conduct two main tasks:
(1) learn the causal factors from the input features with an encoder $\Psi(\cdot)$; (2) provide a certifiable guarantee for robustness in this process.


\looseness=-1
For the first task, encouraged by recent progress in causal generalization, we extract the causal factors of input features in the latent space through techniques in invariant learning \cite{krueger2021out, ahuja2020invariant, arjovsky2019invariant, mitrovic2020representation}, which capture invariant factors across different domains. 
We can adopt one of the cutting-edge methods of this type for our invariant learning module. In this work, we leverage one of the most representative methods: invariant risk minimization (IRM) \cite{arjovsky2019invariant} with the following optimization loss: 


\begin{equation}
\mathcal{L}_{\mathrm{IRM}}=\sum_{d\in\mathbb{D}_{\mathrm{tr}}}R^d(\beta\circ\Psi)+\lambda\cdot\|\nabla_{w|w=1.0}R^{d}(w(\beta\circ\Psi))\|^{2},
\label{eq:IRM}
\end{equation}
\looseness=-1
where $R^d(\beta\circ\Psi)=\mathbb{E}[\mathcal{L}(g(\Psi(x)), y)]$ is the prediction loss in domain $d$ with an encoder $\Psi$ and classifier $\beta$. $w$ is a ``dummy" classifier and can be fixed as a scalar 1.0. According to \cite{arjovsky2019invariant}, the gradient of $R^{d}(w(\beta\circ\Psi))$ reflects the invariance of the learned latent representations. The non-negative hyperparameter $\lambda$ controls the balance between the predictive ability and invariance. 

\looseness=-1
Even though causal factor learning usually does not have specific restrictions regarding the encoder architecture, it is worth noting that an arbitrary architecture cannot provide certifiable robustness in the latent space. 
Therefore, for the second task, we adopt the 1-Lipschitz network \cite{trockman2021orthogonalizing} to derive certifiable robustness across domains.

\subsection{Certified Robustness for Unseen Domains}
While significant progress has been made in certified defenses when training and test data share the same distribution, there is still limited exploration and a lack of theoretical guarantees for certified robustness under domain shifts. In this subsection, we bridge this gap by utilizing the theoretical support from certified defense \cite{cohen2019certified}
and causal inference to derive necessary theorems in this setting. 
According to previous discussions, we perform random smoothing for the causal factors in the latent space. Therefore, based on the calculation of the certified radius, we introduce the following Theorem 1: \vspace{\baselineskip}

\noindent\textbf{Theorem 1.}
\begin{itshape}
Suppose we have an causal encoder $\Psi: \mathcal{X}\rightarrow\mathcal{Z}$, and an arbitrary classifier $\beta: \mathcal{Z}\rightarrow\mathcal{Y}$. Let $g_\beta$ be defined as $g_\beta(z)=\underset{y\in\mathcal{Y}}{\arg\max}P(\beta(z+\eta)=y)$, where $\eta\sim\mathcal{N}(0, \sigma^2I)$, $z=\Psi(x)$ is the latent causal representation. Suppose $\underline{p_A}$ is the lower bound of $p_A$,  $\overline{p_B}$ is the upper bound of $p_B$, $\underline{p_A}, \overline{p_B}\in[0,1]$ and satisfy:
\begin{equation}
P(\beta(z+\eta)=y_A)\ge\underline{p_A}\ge\overline{p_B}\ge\max_{y\ne y_A}P(\beta(z+\eta)=y).
\end{equation}
Then $\forall d,d^{\prime} \in \mathbb{D}$, $g_\beta(z^d+\delta_z)=g_\beta(z^{d^{\prime}}+\delta_z)=y_A$ for all $\|\delta_z\|_2<CR_z$, where $\delta_z$ is the perturbation applied to latent causal representation $z$ and
\begin{equation}
    CR_z(\beta; x, y)=\frac{\sigma}{2}(\Phi^{-1}(\underline{p_A})-\Phi^{-1}(\overline{p_B})).
\end{equation}
\end{itshape}

\looseness=-1
\noindent This theorem provides us a theoretical guarantee that any perturbation $\delta_z$ within the range $CR_z$ will not change the prediction of the smoothed classifier $g_\beta$. It also provides a theoretical guarantee for generalization: for any two instances from $d$ and $d'$ respectively, if their causal latent representations (denoted by $z^d$ and $z^{d'}$) learned from the causal encoder $\Psi$ are the same, then the predictions of $g_\beta$ for them are consistent. Moreover, the certified radius for $z$ across these domains will also be consistent. Therefore, Theorem 1 provides theoretical support for achieving certified robustness on data in different domains by performing random smoothing in the latent space.

Another significant problem left is that the certified radius $CR_z$ mentioned in Theorem 1 is obtained by applying Gaussian noise within latent space and then performing Monte Carlo sampling. Thus, the robustness guarantee is only for $\beta$. However, in practice, attackers often directly perturb input features. Therefore, the certified radius obtained in the latent space needs to be mapped back to the input space to provide certified robustness for the entire classifier $f$. Correspondingly, we have Theorem 2 as follows:\vspace{\baselineskip}

\noindent\textbf{Theorem 2.} 
\begin{itshape}
Let the causal encoder $\Psi$ be $L$-Lipschitz. Let $g$ be defined as $g(x)=\underset{y\in\mathcal{Y}}{\arg\max}P(\beta(\Psi(x+\eta))=y)$. Then $g(x^d+\delta)=g(x^{d^{\prime}}+\delta)=y_A$ for all $\|\delta\|_2<CR_z/L$, where $\delta$ is the perturbation applied to input features $x$.  
\end{itshape}\vspace{\baselineskip}

\looseness=-1
\noindent Briefly, if we use an $L$-Lipschitz neural network in the causal factor learning module, we can calculate the certified radius in the input space. This is because we can simply scale the certified radius in the latent space by the Lipschitz constant $L$, such that $CR\geq{CR_z/L}$. If $L=1$, then $CR_z$ will be the lower bound of $CR$. With the aforementioned causal invariant assumption, the certified robustness for instances in one domain can also be propagated to instances in other domains with the same causal factors. 
Therefore, we are able to provide theoretical guarantees for cross-domain certified robustness. Detailed proofs of Theorem 1 and Theorem 2 can be found in the Appendix. 

\subsection{Implementation}
\looseness=-1
\subsubsection{Overview of Framework}
We integrate the previous methods and theories to form our framework, which is demonstrated in Figure \ref{fig2}. In Figure \ref{fig2}, the gray path represents the training process. During training, we apply Gaussian augmentation to $z$ to enhance the prediction accuracy during the RS phase. The green path represents the certifying process. We first train the causal encoder $\Psi$ and classifier $\beta$, then obtain robustness guarantees for the classifier $\beta$ by adding Gaussian noise to $z$ with Monte Carlo sampling. The bottom path represents the mapping process. Specifically, it involves multiplying the certified radius in the latent space by the mapping constant $1/L$, and reverting back to the input space to obtain robustness guarantees for the input feature $x$. 

\subsubsection{Architecture}
As aforementioned, we use Lipschitz constraints in the causal factor learning module. We define the final linear layer as the classifier $\beta$, with all preceding layers forming the encoder $\Psi$. We apply the Cayley transform \cite{trockman2021orthogonalizing} to achieve orthogonality, thereby ensuring that each linear layer has a Lipschitz constant of 1. For the activation functions, we employ GroupSort \cite{anil2019sorting}, which also has 1-Lipschitzness. More details on the implementation of 1-Lipschitz networks can be found in the Appendix.

\begin{table*}[ht]
\centering
\caption{A comparison of certified test accuracy (\%) and ACR between our framework and baselines. For each method, we recorded data for ten radii $r$ ranging from 0.00 to 0.45, with increments of 0.05. Every model is certified with $\sigma=0.12$. We highlight our results in bold whenever the value improves the baselines.}
\begin{tabular}{ccccccccccccc}
\hline
Datasets                   & Models        & $r=0.00$          & 0.05          & 0.10          & 0.15          & 0.20          & 0.25          & 0.30          & 0.35          & 0.40          & 0.45          & ACR             \\ \hline
\multirow{5}{*}{CMNIST}    & Gaussian      & 18.1          & 14.8          & 12.9          & 10.5          & 9.0           & 8.6           & 8.1           & 7.8           & 7.3           & 6.0           & 0.0458          \\
                           & MACER         & 23.6          & 19.4          & 15.5          & 12.4          & 10.1          & 8.8           & 7.4           & 5.6           & 4.1           & 1.9           & 0.0482          \\
                           & SmoothAdv     & 27.2          & 22.7          & 16.9          & 13.6          & 10.5          & 8.5           & 7.3           & 5.8           & 3.6           & 1.3           & 0.0518          \\
                           & Consistency   & 12.1          & 11.3          & 10.8          & 10.7          & 10.6          & 10.5          & 10.4          & 10.4          & 10.3          & 10.3          & 0.0488          \\
                           & \textbf{\mymodel~(Ours)} & \textbf{64.3} & \textbf{62.7} & \textbf{60.5} & \textbf{58.0} & \textbf{55.8} & \textbf{53.2} & \textbf{51.4} & \textbf{47.9} & \textbf{44.9} & \textbf{38.6} & \textbf{0.2466} \\ \hline
\multirow{5}{*}{CelebA}    & Gaussian      & 31.0          & 31.0          & 31.0          & 29.0          & 28.0          & 26.0          & 26.0          & 24.0          & 22.0          & 17.0          & 0.1218          \\
                           & MACER         & 21.0          & 19.0          & 15.0          & 13.0          & 10.0          & 10.0          & 10.0          & 8.0           & 7.0           & 6.0           & 0.053           \\
                           & SmoothAdv     & 25.0          & 23.0          & 18.0          & 16.0          & 14.0          & 11.0          & 10.0          & 9.0           & 9.0           & 5.0           & 0.0623          \\
                           & Consistency   & 24.0          & 24.0          & 23.0          & 23.0          & 22.0          & 21.0          & 20.0          & 20.0          & 20.0          & 20.0          & 0.0989          \\
                           & \textbf{\mymodel~(Ours)} & \textbf{62.0} & \textbf{59.0} & \textbf{57.0} & \textbf{56.0} & \textbf{52.0} & \textbf{51.0} & \textbf{49.0} & \textbf{45.0} & \textbf{42.0} & \textbf{38.0} & \textbf{0.2326} \\ \hline
\multirow{5}{*}{DomainNet} & Gaussian      & 33.0          & 32.0          & 32.0          & 31.0          & 28.0          & 25.0          & 24.0          & 23.0          & 22.0          & 19.0          & 0.1223          \\
                           & MACER         & 28.0          & 28.0          & 28.0          & 28.0          & 28.0          & 28.0          & 28.0          & 28.0          & 27.0          & 27.0          & 0.1272          \\
                           & SmoothAdv     & 27.0          & 27.0          & 26.0          & 26.0          & 25.0          & 24.0          & 23.0          & 23.0          & 20.0          & 18.0          & 0.1097          \\
                           & Consistency   & 27.0          & 27.0          & 27.0          & 27.0          & 27.0          & 27.0          & 27.0          & 27.0          & 27.0          & 27.0          & 0.1234          \\
                           & \textbf{\mymodel~(Ours)} & \textbf{63.0} & \textbf{61.0} & \textbf{55.0} & \textbf{54.0} & \textbf{48.0} & \textbf{43.0} & \textbf{41.0} & \textbf{37.0} & \textbf{30.0} & 23.0          & \textbf{0.2088} \\ \hline
\end{tabular}
\label{table 2}
\end{table*}

\section{EXPERIMENTS}



\looseness=-1
In this section, we conduct extensive experiments to evaluate our framework on one synthetic dataset and two real-world datasets. Specifically, we answer the following research question based on the experimental results. \textbf{RQ1}: How does \mymodel~perform compared to the baselines of certified defense? \textbf{RQ2}: How do different components in \mymodel~contribute to the performance? \textbf{RQ3}: How does \mymodel~perform under different settings of hyperparameters?

\subsection{Datasets}
We introduce the three datasets used in the experiments: CMNIST \cite{arjovsky2019invariant}, CelebA \cite{liu2015faceattributes} and DomainNet \cite{peng2019moment}. Detailed information on the domain construction and division of all these three datasets can be found in the Appendix.

\subsection{Experiment Settings}
\textbf{Baselines.} We evaluate our framework by comparing it against several representative certified defense methods. All these methods are based on RS: 
\begin{itemize}
    \item\textbf{Gaussian} \cite{cohen2019certified}: Standard training with Gaussian noise based random smoothing.
    \item\textbf{MACER} \cite{zhai2020macer}: Add a regularization term that maximizes an approximate form of the certified radius.
    \item\textbf{SmoothAdv} \cite{salman2019provably}: Adversarial training is incorporated during the training of the smoothed classifier.
    \item\textbf{Consistency} \cite{jeong2020consistency}: The Kullback-Leibler divergence between the mean of the classifier's predictions after various perturbations and the prediction after a single perturbation was used as a regularization term. This term minimizes the variance in the classifier's predictions after different perturbations, optimizing the objective for robust training of the smoothed classifier.
\end{itemize}

\looseness=-1
\noindent\textbf{Evaluation Metrics.} We consider two widely-used evaluation metrics: (1) \textit{certified accuracy} at different \textit{radii}, which is defined as the fraction of the test set that CERTIFY \cite{cohen2019certified} classifies correctly. CERTIFY is a practical Monte Carlo-based certification procedure that offers the prediction of $g$ along with the lower bound of the certified radius or abstains the certification by sampling over $n$ Gaussian noises with the probability of at least $1-\alpha$, $\alpha$ is the significance level; (2) \textit{average certified radius} (ACR), which is defined as $ACR=\frac{1}{|\mathbb{D}_{\mathtt{test}}|}\sum_{(x,y)\in\mathbb{D}_{\mathtt{test}}}{CR}(f;\sigma,x,y)\cdot\mathbf{1}_{[g(x,\sigma)=y]}$. Here, $|\mathbb{D}_{\mathtt{test}}|$ is the capacity of the test set, $CR$ is the certified radius returned by CERTIFY, $\mathbf{1}$ is the indicator function. We assign 0 to $CR$ for incorrect prediction of $g$. We use the same settings in \cite{cohen2019certified} with $n=100000, n_0=100, \alpha=0.001$ to apply CERFITY. Here $n_0$ is the small number of samples to find $y_A$. Note that, for two different models, their certified accuracies sometimes cannot be directly compared. 
At a specific radius $r$, one model may have a higher certified accuracy than the other, but the situation may be reversed at another radius. Therefore, ACR is a more suitable metric as it reflects ``average robustness".

\noindent\textbf{Training Details.} We use a three-layer MLP for CMNIST and a four-layer CNN for CelebA and DomainNet. During inference, we apply RS with the noise level $\sigma=0.12$. The result of other $\sigma$ is shown in Appendix. We set the parameter of the regularization term $\lambda=10000$ for all datasets.

\subsection{Experiment Results}

\begin{figure*}[htbp]
\centering
\begin{subfigure}[]{0.3\linewidth}
    \includegraphics[width=\linewidth]{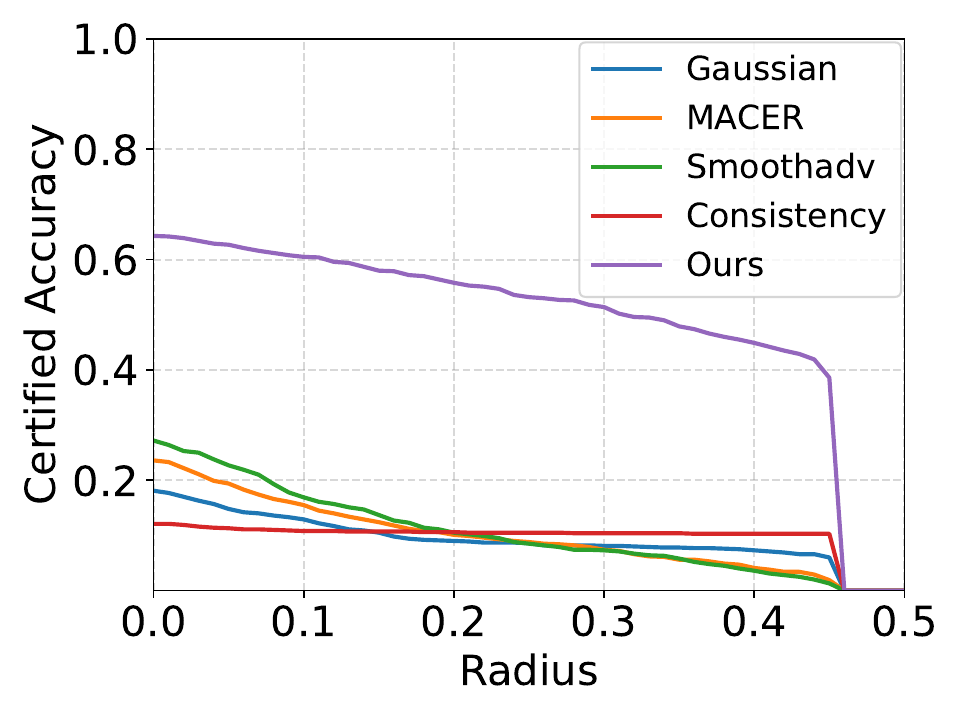}
    \caption{CMNIST}
    \label{fig4_1}
\end{subfigure}
\hfill
\begin{subfigure}[]{0.3\linewidth}
    \includegraphics[width=\linewidth]{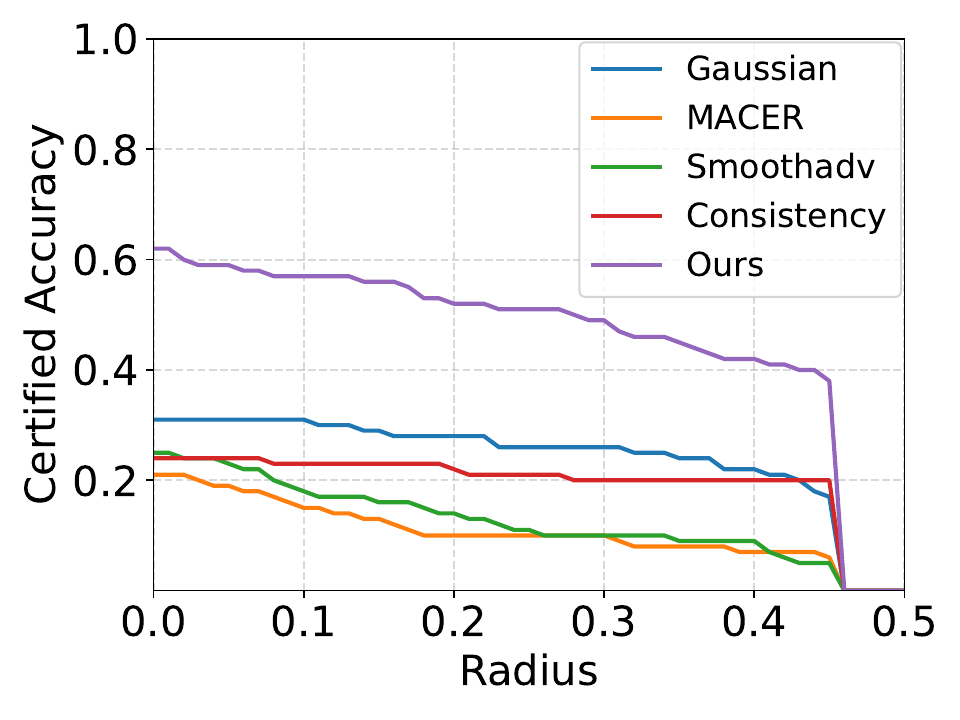}
    \caption{CelebA}
    \label{fig4_2}
\end{subfigure}
\hfill
\begin{subfigure}[]{0.3\linewidth}
    \includegraphics[width=\linewidth]{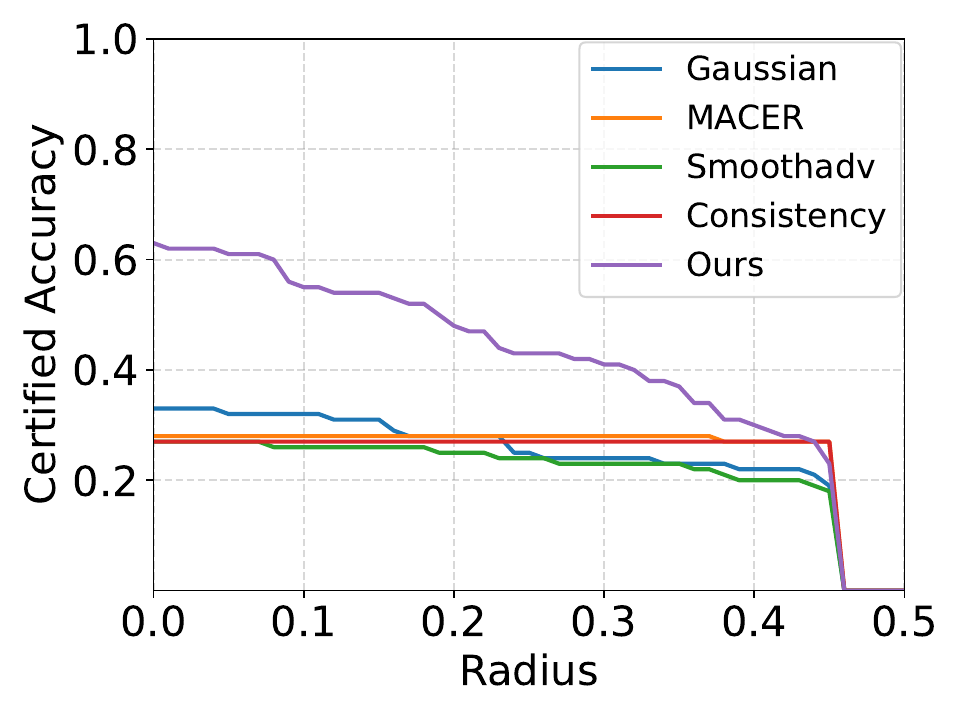}
    \caption{DomainNet}
    \label{fig4_3}
\end{subfigure}
\caption{Comparison of certified accuracy obtained using different methods across three datasets. The sharp decline at the end of the curves is due to a hard upper bound in the certification process for a variance $\sigma$ and the number of Gaussian samples $n$. }
\vspace{-3mm}
\label{fig3}
\end{figure*}

\subsubsection{Performance}
\looseness=-1
For all datasets, more detailed settings for the training parameters are provided in the Appendix. Table \ref{table 2} shows a comparison of performance between our framework and baselines w.r.t. ACR and the certified test accuracy with different radii $r$. We also plot the radius-certified accuracy curve in Figure \ref{fig3}. Note that ACR is equivalent to the area under the curve. From the results, we observe that our method achieves the highest certified accuracy and ACR (with a significant improvement compared with others) at almost all radii across the three datasets. 
Given that our training and testing data reside in different domains, the experimental results demonstrate that our approach significantly and consistently outperforms baselines in the generalization of certified robustness across domains. We omit the variance of the experimental results because it is far smaller than the performance gap between the methods. From Table 2, we can also observe that the ACR decreases progressively from CMNIST to CelebA to DomainNet. This decline is reasonable because the CMNIST dataset only involves spurious correlations between color and digits, whereas CelebA, in addition to the constructed spurious correlation between smiling and hair color, includes more complicated domain shifts regarding other facial features. For DomainNet, the complex variation in backgrounds makes the causal relationships within the data more difficult to capture.

\begin{figure}[ht]
\centering
\begin{subfigure}[b]{0.45\linewidth}
    \includegraphics[width=\linewidth]{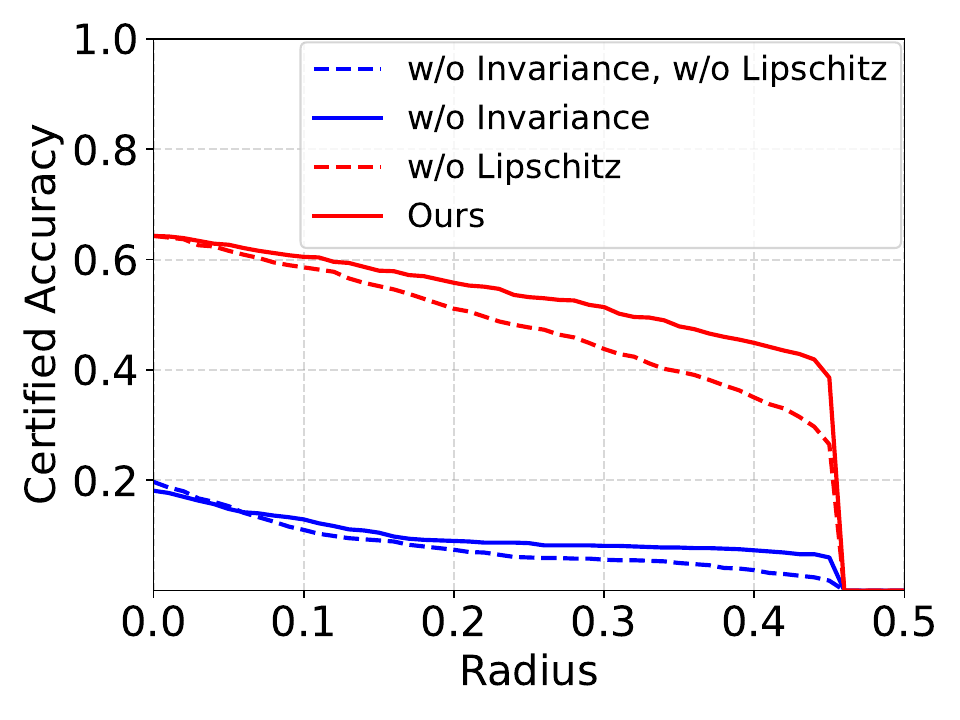}
    \vspace{-3mm}  
    \caption{$\sigma=0.12$}
    \label{Figure4_1}
\end{subfigure}
\hfill 
\begin{subfigure}[b]{0.45\linewidth}
    \includegraphics[width=\linewidth]{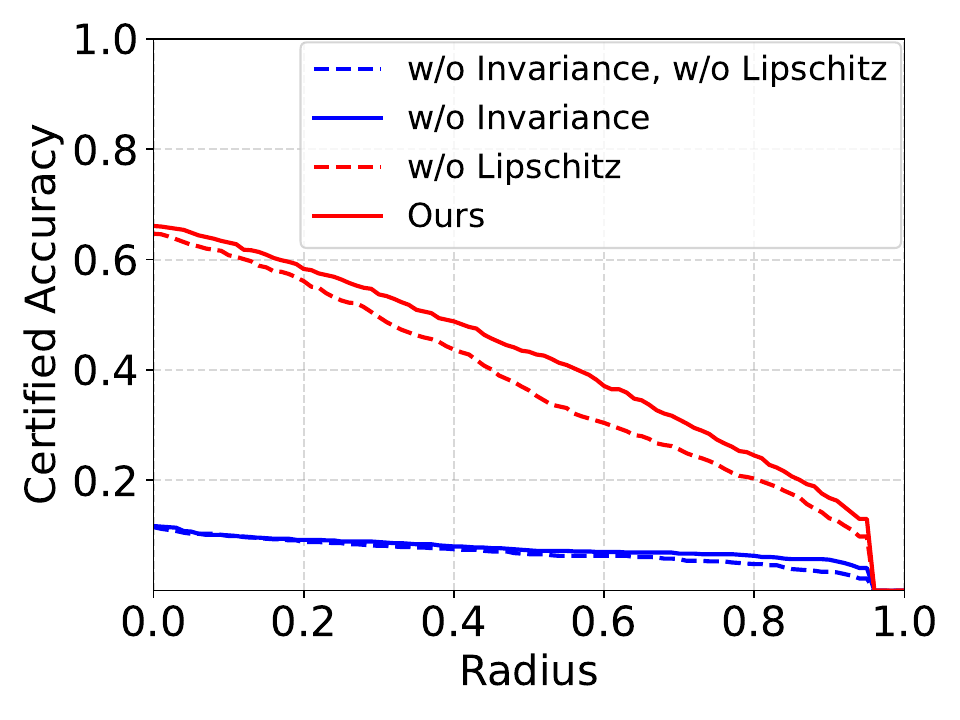}
    \vspace{-3mm}  
    \caption{$\sigma=0.25$}
    \label{Figure4_2}
\end{subfigure}

\caption{Ablation study on CMNIST with different $\sigma$.}
\vspace{-3mm}
\label{Figure4}
\end{figure}

\looseness=-1
\subsubsection{Ablation Study} To evaluate the effectiveness of each component in our method, we
provide ablation study with the following variants: (1) w/o invariance: We remove the invariance regularization term in Eq.(\ref{eq:IRM}) and only use the first term as an ERM loss. (2) Network without Lipschitz Constraints: We replace the 1-Lipschitz layers in the network with the ones without any constraints. We conducted comparisons with two types of ablation studies simultaneously under $\sigma=0.12$ and $\sigma=0.25$. As shown in Figure \ref{Figure4}, our model undoubtedly outperforms the version without the invariant penalty since this variant cannot capture the causal factors effectively and thus fails to mitigate the influence of spurious correlations on robustness. For our model, the certified accuracy and ACR with Lipschitz constraints is slightly better than that of networks without any constraints. This is because Lipschitz constraints ensure that we use causal factors for certification. 
The results of the other two datasets are provided in the Appendix.

\begin{figure}[ht]
\centering
\begin{subfigure}[b]{0.45\linewidth}
    \includegraphics[width=\linewidth]{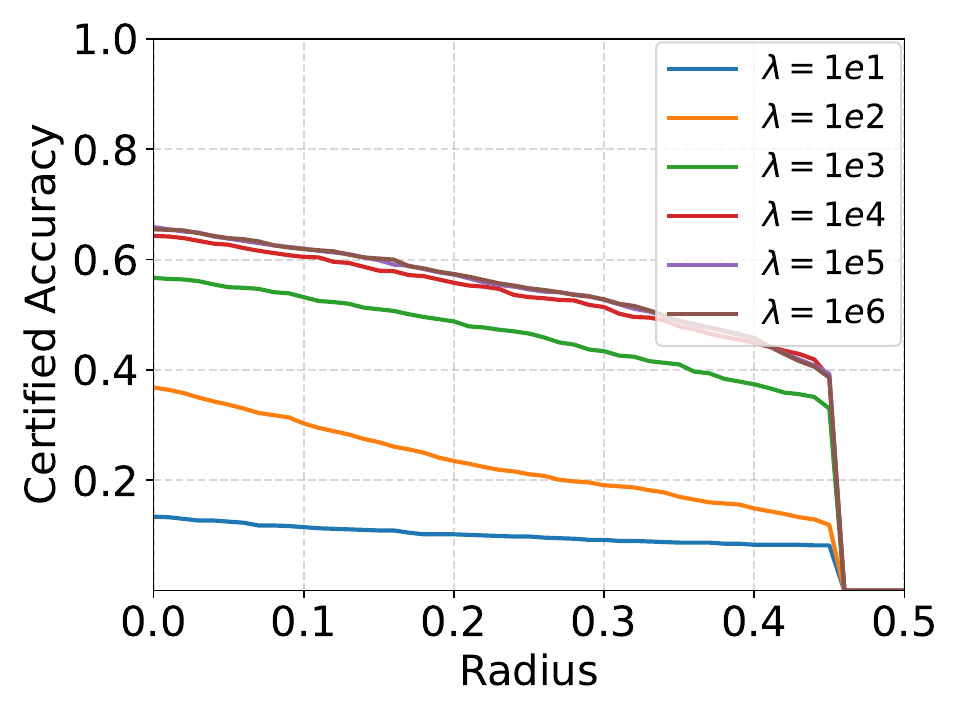}
    \vspace{-3mm}  
    \caption{Vary $\lambda$}
    \label{Figure5_1}
\end{subfigure}
\hfill 
\begin{subfigure}[b]{0.45\linewidth}
    \includegraphics[width=\linewidth]{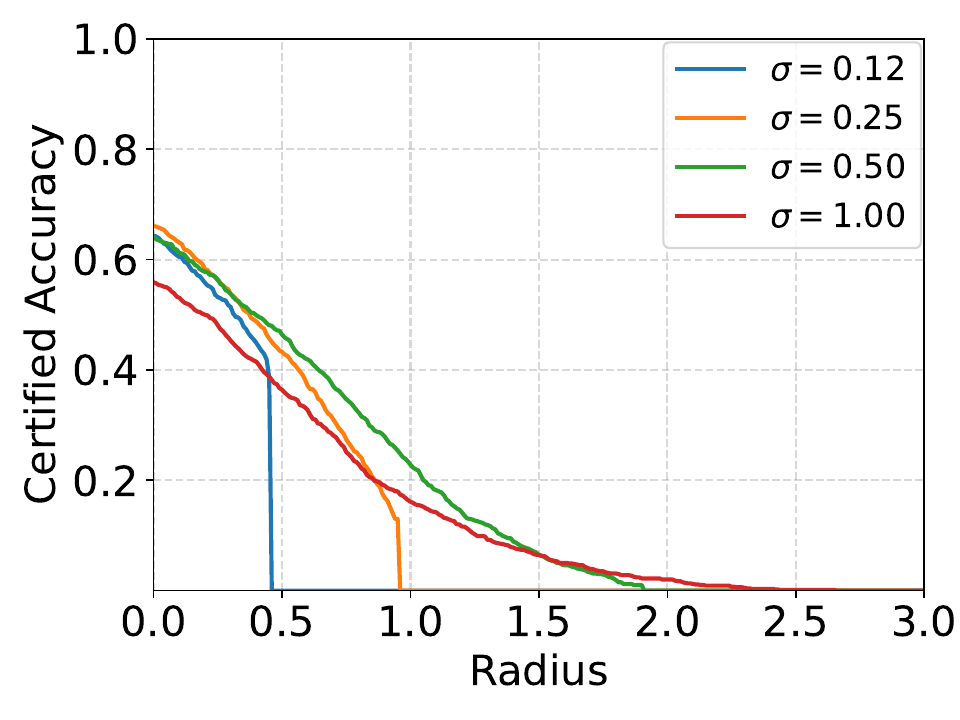}
    \vspace{-3mm}  
    \caption{Vary $\sigma$}
    \label{Figure5_2}
\end{subfigure}

\caption{Performance of \mymodel~under different parameters.}
\vspace{-3mm}
\label{Figure5}
\end{figure}




\looseness=-1
\subsubsection{Parameter Study} We set the hyperparameter $\lambda\in\{10, 10^2, 10^3, 10^4, 10^5, 10^6\}$, $\sigma\in\{0.25, 0.50, 1.00\}$. The results of the parameter study on CMNIST are shown in Figure \ref{Figure5}. The results for the other two datasets are provided in the Appendix. We can observe in Figure \ref{Figure5} (a) that when $\lambda$ increases, the certified accuracy at the same radius also increases, this is because a higher $\lambda$ leads to stronger causal factor learning, and achieving stronger generalizable robustness. However, when $\lambda$ exceeds 10,000, the improvement in model performance becomes negligible as it has reached the bottleneck of the model's ability to learn invariant causal factors. As shown in Figure \ref{Figure5} (b) $\sigma$ controls the level of noise. A higher noise level means that we can obtain a larger certified radius but at the cost of reduced certified accuracy. 

\vspace{-2mm}
\section{Conclusion}
In this paper, we address the critical problem of generalizing certified robustness across different domains. We analyze the limitations of existing certified defense strategies and explore the challenges posed by robustness under domain shifts. To address this problem, we introduce a novel causality-inspired framework, \mymodel, designed to learn causal factors that mitigate the negative impact of spurious correlations on robustness, enabling a certifiable defense process across various domains. Extensive experiments on both synthetic and real-world benchmarks verify the effectiveness of our method. \mymodel~can pave the path for future work that aims at further exploring causality-inspired defenses and any unified approaches for the generalization of adversarial robustness.

\bibliography{aaai25}
\clearpage
\appendix
\section{Appendix}

\subsection{Theoretical Analysis}
In this section, we provide detailed proofs of the two theorems proposed in our work.

\subsubsection{Proof of Theorem 1} Our proof is based on Theorem 1 of \cite{cohen2019certified}, which is summarized in our Preliminaries and Related Work section with Eq. \ref{eq3}. But differently, we extend the theory by (1) deriving the certified radius through random smoothing in the latent space; and (2) generalizing it to different data domains. Suppose we have a causal encoder $\Psi:\mathcal{X}\rightarrow\mathcal{Z}$, and a classifier $\beta:\mathcal{Z}\rightarrow\mathcal{Y}$, we can formally define $g_\beta$, i.e., the smoothed version of  $\beta$ as:
\begin{equation}
    g_\beta(z)=\underset{y\in\mathcal{Y}}{\arg\max}P(\beta(z+\eta)=y),
\end{equation}
where $z=\Psi(x)$ is the latent causal representation. Based on Eq. \eqref{eq3}, we can determine the certified radius in latent space:
\begin{equation}
\label{eq12}
        CR_z(\beta; x, y)=\frac{\sigma}{2}(\Phi^{-1}(p_A)-\Phi^{-1}(p_B)),
\end{equation}
but here in \eqref{eq12}, $p_A=P(\beta(z+\eta)=y_A), p_B=\max_{y\ne y_A}P(\beta(z+\eta)=y)$. Then any perturbation $\delta_z$ on $z$ within radius $CR_z$ will not change the prediction of $g_\beta$.

Next, we prove that under the condition $\|\delta_z\|_2\leq{CR_z}$, the prediction of $g_\beta$ remains unchanged across different domains. Recall that:
\begin{equation}
\label{eq13}
    P^{d}(Y|C) = P^{d^{\prime}}(Y|C),\forall d,d^{\prime} \in \mathbb{D}.
\end{equation}
For any two instances $x^d$ and $x^{d^\prime}$ in different domains but have the same prediction $y$, we denote their corresponding causal representations by $z^d$ and $z^{d^\prime}$. Then based on \eqref{eq13}, $\beta(z^d)=\beta(z^{d^{\prime}})$. Considering the same Gaussian noise $\eta\sim\mathcal{N}(0, \sigma^2I)$, suppose $\eta$ cannot change the distribution of $\beta(z+\eta)$, then we have:
\begin{equation}
P(\beta(z^d+\eta)=p_A)=P(\beta(z^{d^{\prime}}+\eta)=p_A).
\end{equation}
Therefore, $\forall d,d^{\prime} \in \mathbb{D}$, $g_\beta(z^d+\delta_z)=g_\beta(z^{d^{\prime}}+\delta_z)=y_A$.

\subsubsection{Proof of Theorem 2} Given an L-Lipschitz encoder $\Psi$, for any input instance $x$, we have:
\begin{equation}
\label{eq15}
    \|z-z^\prime\|_2=\|\Psi(x)-\Psi(x^\prime)\|_2\leq{L}\|x-x^\prime\|.
\end{equation}
Let $\|\delta_z\|_2=\|z-z^\prime\|_2$, $\|\delta\|_2=\|x-x^\prime\|_2$, then according to \eqref{eq15}:
\begin{equation}
\label{ieq16}
    \|\delta_z\|_2\leq{L}\|\delta\|_2. 
\end{equation}
Recall Theorem 1 that: if $\|\delta_z\|_2\leq{CR_z}$, then $\forall d,d^{\prime} \in \mathbb{D}$,
\begin{equation}
g_\beta(z^d+\delta_z)=g_\beta(z^{d^{\prime}}+\delta_z)=y_A. 
\end{equation}
Our target is to derive the certified radius in the input space, denoted by $CR$, which should satisfy that for any
$\|\delta\|_2\leq{CR}$, $\forall d,d^{\prime} \in \mathbb{D}$, 
\begin{equation}
\label{eq18}
g(x^d+\delta)=g(x^{d^{\prime}}+\delta)=y_A.
\end{equation}
If $\|\delta\|_2\leq{CR}$, then $L\|\delta\|_2\leq{L\cdot{CR}}$. We have the inequality \eqref{ieq16}, and with the addition of $\|\delta_z\|_2\leq{CR_z}$ and $L\|\delta\|_2\leq{L\cdot{CR}}$, we can derive the condition that satisfies the proposition \eqref{eq18}:
\begin{equation}
    CR\ge{CR_z}/L,
\end{equation}
which means $CR_z/L$ is the lower bound of the certified radius in the input space.

\begin{table}[]
\centering
\caption{Domain split for CMNIST and the number of examples in each group.}
\label{tab3}
\begin{tabular}{cccc}
\hline
Domain                      & Class        & Red   & Green \\ \hline
\multirow{2}{*}{Training 1} & Small Digits & 2500  & 10000 \\
                            & Large Digits & 10000 & 2500  \\ \hline
\multirow{2}{*}{Training 2} & Small Digits & 1250  & 11250 \\
                            & Large Digits & 11250 & 1250  \\ \hline
\multirow{2}{*}{Test}       & Small Digits & 4500  & 500   \\
                            & Large Digits & 500   & 4500  \\ \hline
\end{tabular}
\vspace{-3mm}
\end{table}

\begin{table}[]
\centering
\caption{Domain split for CelebA and the number of examples in each group.}
\label{tab4}
\begin{tabular}{cccc}
\hline
Domain                      & Class      & Blond & Not Blond \\ \hline
\multirow{2}{*}{Training 1} & Smiling    & 500   & 4500      \\
                            & No Smiling & 4500  & 500       \\ \hline
\multirow{2}{*}{Training 2} & Smiling    & 1000  & 4000      \\
                            & No Smiling & 4000  & 1000      \\ \hline
\multirow{2}{*}{Test}       & Smiling    & 1800  & 200       \\
                            & No Smiling & 200   & 1800      \\ \hline
\end{tabular}
\vspace{-3mm}
\end{table}

\begin{table}[t]
\centering
\caption{Domain split for DomainNet and the number of examples in each group.}
\label{tab5}
\begin{tabular}{cccc}
\hline
Domain                      & Class   & Clipart & Sketch \\ \hline
\multirow{2}{*}{Training 1} & Animal  & 450     & 50     \\
                            & Vehicle & 50      & 450    \\ \hline
\multirow{2}{*}{Training 2} & Animal  & 400     & 100    \\
                            & Vehicle & 100     & 400    \\ \hline
\multirow{2}{*}{Test}       & Animal  & 10      & 90     \\
                            & Vehicle & 90      & 10     \\ \hline
\end{tabular}
\vspace{-3mm}
\end{table}

\begin{table*}[ht]
\centering
\caption{Comparison of certified test accuracy (\%) and ACR between our framework and baselines. For each method, we recorded data for ten radii $r$ ranging from 0.00 to 0.90, with increments of 0.10. Every model is certified with $\sigma = 0.25$. We highlight our results in bold whenever the value improves the baselines.}
\label{tab6}
\begin{tabular}{ccccccccccccc}
\hline
Datasets                   & Models      &$r=$ 0.00          & 0.10          & 0.20          & 0.30          & 0.40          & 0.50          & 0.60          & 0.70          & 0.80          & 0.90          & ACR             \\ \hline
\multirow{5}{*}{CMNIST}    & Gaussian    & 11.7          & 9.9           & 9.2           & 8.8           & 8.0           & 7.3           & 7.0           & 6.7           & 6.3           & 5.6           & 0.0744          \\
                           & MACER       & 11.2          & 10.9          & 10.6          & 10.6          & 10.2          & 9.7           & 9.3           & 8.5           & 7.5           & 5.2           & 0.0833          \\
                           & SmoothAdv   & 17.7          & 13.8          & 10.9          & 9.2           & 8.2           & 7.0           & 6.4           & 4.8           & 2.6           & 0.7           & 0.0727          \\
                           & Consistency & 10.7          & 10.7          & 10.6          & 10.5          & 10.5          & 10.5          & 10.5          & 10.4          & 10.3          & 10.1          & 0.0997          \\
                           & Ours        & \textbf{66.1} & \textbf{63.1} & \textbf{58.3} & \textbf{53.7} & \textbf{48.8} & \textbf{43.3} & \textbf{37.1} & \textbf{31.0} & \textbf{24.5} & \textbf{16.8} & \textbf{0.4100} \\ \hline
\multirow{5}{*}{CelebA}    & Gaussian    & 32.0          & 30.0          & 28.0          & 27.0          & 24.0          & 20.0          & 15.0          & 14.0          & 14.0          & 13.0          & 0.1999          \\
                           & MACER       & 23.0          & 22.0          & 20.0          & 16.0          & 15.0          & 12.0          & 10.0          & 10.0          & 9.0           & 9.0           & 0.1337          \\
                           & SmoothAdv   & 29.0          & 22.0          & 18.0          & 12.0          & 9.0           & 9.0           & 8.0           & 7.0           & 6.0           & 4.0           & 0.1085          \\
                           & Consistency & 24.0          & 24.0          & 22.0          & 20.0          & 20.0          & 20.0          & 20.0          & 18.0          & 18.0          & 17.0          & 0.1923          \\
                           & Ours        & \textbf{56.0} & \textbf{54.0} & \textbf{52.0} & \textbf{49.0} & \textbf{43.0} & \textbf{38.0} & \textbf{33.0} & \textbf{31.0} & \textbf{26.0} & \textbf{21.0} & \textbf{0.3772} \\ \hline
\multirow{5}{*}{DomainNet} & Gaussian    & 32.0          & 30.0          & 27.0          & 25.0          & 24.0          & 24.0          & 18.0          & 18.0          & 15.0          & 11.0          & 0.2108          \\
                           & MACER       & 27.0          & 27.0          & 27.0          & 27.0          & 26.0          & 26.0          & 25.0          & 24.0          & 24.0          & 24.0          & 0.2440          \\
                           & SmoothAdv   & 33.0          & 29.0          & 25.0          & 23.0          & 20.0          & 18.0          & 15.0          & 12.0          & 12.0          & 10.0          & 0.1816          \\
                           & Consistency & 26.0          & 26.0          & 26.0          & 26.0          & 25.0          & 25.0          & 25.0          & 25.0          & 25.0          & 24.0          & 0.2413          \\
                           & Ours        & \textbf{53.0} & \textbf{48.0} & \textbf{43.0} & \textbf{37.0} & \textbf{31.0} & 22.0          & 19.0          & 18.0          & 15.0          & 13.0          & \textbf{0.2749} \\ \hline
\end{tabular}
\vspace{-3mm}
\end{table*}

\begin{figure*}[htbp]
\centering
\begin{subfigure}[]{0.3\linewidth}
    \includegraphics[width=\linewidth]{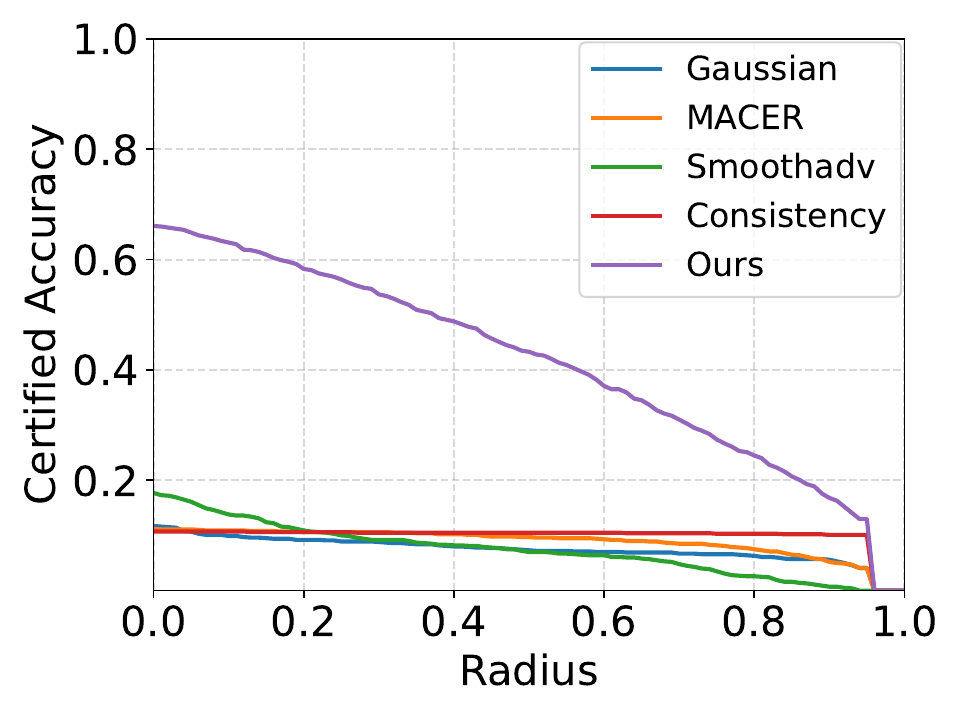}
    \caption{CMNIST}
    \label{fig6_1}
\end{subfigure}
\hfill
\begin{subfigure}[]{0.3\linewidth}
    \includegraphics[width=\linewidth]{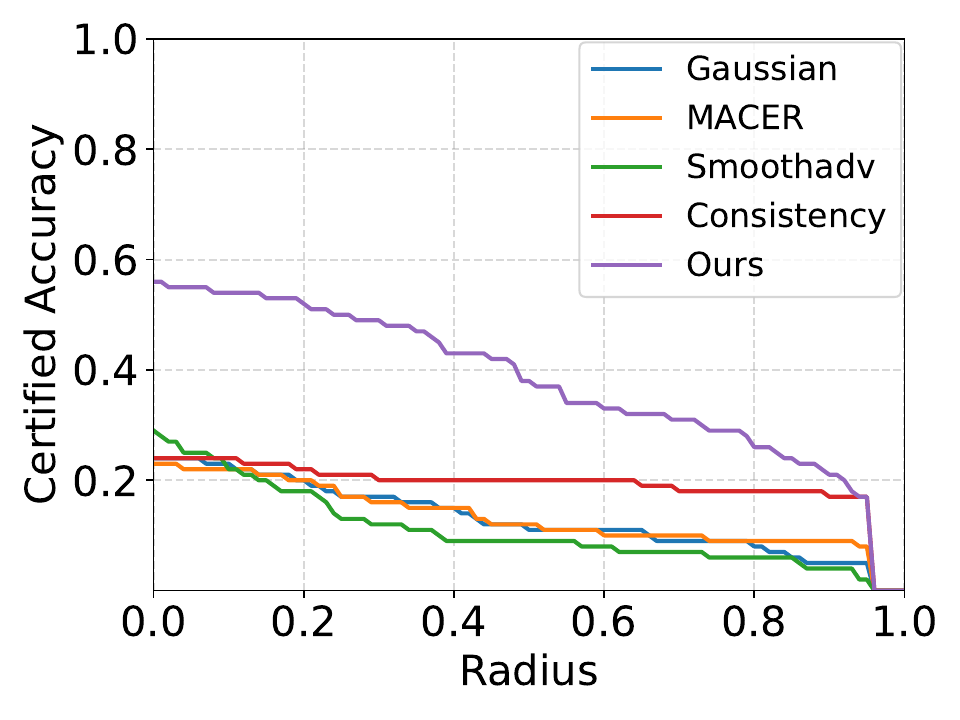}
    \caption{CelebA}
    \label{fig6_2}
\end{subfigure}
\hfill
\begin{subfigure}[]{0.3\linewidth}
    \includegraphics[width=\linewidth]{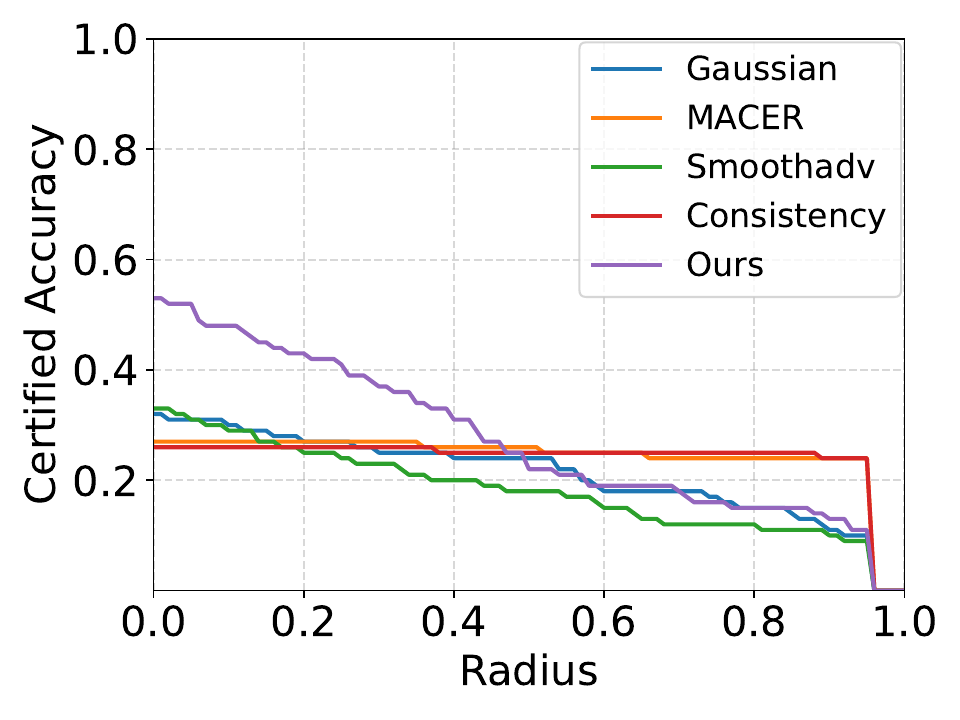}
    \caption{DomainNet}
    \label{fig6_3}
\end{subfigure}
\caption{Comparison of certified accuracy obtained using different methods across three datasets ($\sigma=0.25$). }
\label{fig6}
\end{figure*}

\subsection{Details on Experimental Setups}
\subsubsection{Training Details} We use Ubuntu 22.04 and deploy all models on an NVIDIA GeForce 4090 GPU with 24GB of memory. For CMNIST, every model is trained for 500 epochs ($e=500$) with $\lambda$ set to 1 for the first 100 epochs ($e_0=100$) because we want the model initially learn a more comprehensive semantic understanding of the images. For CelebA, $e$ is set to 50 and $e_0$ to 20, whereas for DomainNet, the values are set to 200 and 30, respectively. For all datasets, we use the Adam optimizer with a global learning rate set at 0.001. For CMNIST, the entire dataset is treated as a single batch. For CelebA and DomainNet, we resize the images to $64\times64$, with the batch size set at 512.

\subsubsection{Datasets} 
Here, we introduce more details of dataset setup.

\noindent
\begin{itemize}
    \item \textbf{CMNIST} \cite{arjovsky2019invariant}: This dataset is a modified version of MNIST. It assigns label 0 to digits 0 through 4 and label 1 to digits 5 through 9, making CMNIST a dataset with two categories. The dataset encompasses three domains, with two used for training and one for testing. The domain shifts arise from variations in the relationship between digit colors and their labels across different domains. The domain splits of CMNIST are shown in Table \ref{tab3}.
    \item\textbf{CelebA} 
\cite{liu2015faceattributes}: Each image in the CelebA dataset is associated with 40 attributes. In our experiments, similar to the setup with CMNIST, we select the binary attribute ``Smiling" as our classification target and establish two training domains and one testing domain. ``Blond Hair" serves as a spurious feature, whose variation leads to correlation shifts across domains. The domain splits of CelebA are shown in Table \ref{tab4}.

\item \textbf{DomainNet} \cite{peng2019moment}: This dataset contains six domains: \textit{Clipart, Infograph, Painting, Quickdraw, Real, and Sketch}. Each domain includes 345 categories of common objects. We select multiple categories to form two superclasses, animals and vehicles, which serve as our classification targets. The styles of the objects, \textit{Clipart} and \textit{sketch}, are designated as spurious features. Changes in the correlations between the styles and objects cause domain shifts. The domain splits of CelebA are shown in Table \ref{tab5}.
\end{itemize}

\subsubsection{1-Lipschitz Network} For linear layers, we can construct orthogonal linear layers using the Cayley transform:
\begin{equation}
    Q=(I-A)(I+A)^{-1}
\end{equation}
where $Q$ is the orthogonal matrix, $A$ is a skew-symmetric matrix that can be constructed by the difference between the weight matrix $W$ and its conjugate transpose.
However, for convolutional layers, computing the matrix inverse involved in the Cayley transform is time-consuming. Therefore, we employ the Fast Fourier Transform (FFT) to perform calculations of Cayley transform \cite{trockman2021orthogonalizing} in the spectral domain. Additionally, for implementing downsampling, we can increase the number of output channels by 4$\times$ to compensate for 2 $\times$ downsampling. Another approach is to use the average pooling since it is also non-expansive \cite{trockman2021orthogonalizing}. If we use ResNet, considering the residual connections, we utilize the following property to maintain the Lipschitz constant \cite{su2022scaling}: for two L-Lipschitz functions $f_1(x)$ and $f_2(x)$, if $f(x)=\alpha{f_1(x)}+(1-\alpha)f_2(x)$, then $f(x)$ is also L-Lipschitz, where $\alpha\in[0,1]$ is a learnable parameter. 

\subsection{Supplementary Results of Experiments}

\subsubsection{Performance} We apply randomized smoothing (RS) with another noise level $\sigma=0.25$. The results are shown in Table \ref{tab6} and Figure \ref{fig6}. For CMNIST and CelebA datasets, we observe that our method achieves the highest certified accuracy and ACR (with a significant improvement compared with others). However, for the DomainNet dataset, a trade-off occurs. Our model focuses more on robustness at smaller radii $r$  due to the following two reasons. First, trade-offs are commonly encountered in certified defense. Second, baselines typically maximize the coefficient of robust regularization term to force robustness across all radii, but these approaches significantly sacrifice certified accuracy at smaller radii. This leads to robustness achieved at all radii but with consistently lower certified accuracy, making these baselines somewhat meaningless.
As a result, our method exhibits higher certified accuracy when $r<0.5$, but it is slightly lower than the baseline at higher $r$ values. Based on our previous description of metrics, when a model's certified accuracy does not consistently exceed that of another model at every radius $r$, it is more rational to use the ACR since it can reflect the average robustness across all radii. It is evident that our method still maintains the highest ACR.

\begin{figure}[ht]
\centering
\begin{subfigure}[b]{0.45\linewidth}
    \includegraphics[width=\linewidth]{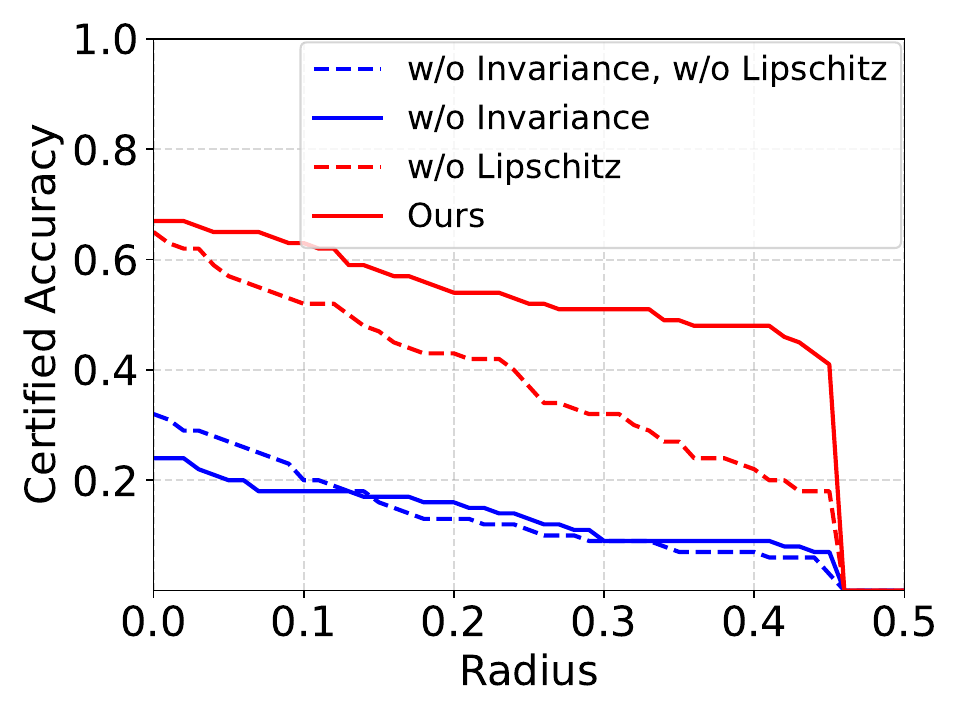}
    \caption{$\sigma=0.12$}
    \label{Figure7_1}
\end{subfigure}
\hfill 
\begin{subfigure}[b]{0.45\linewidth}
    \includegraphics[width=\linewidth]{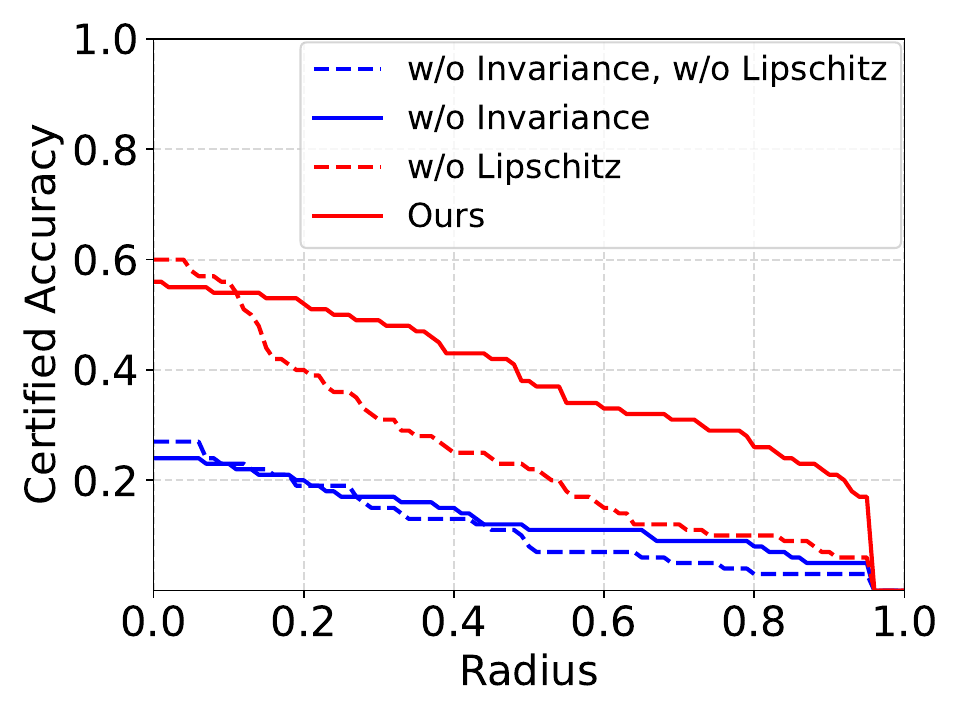}
    \caption{$\sigma=0.25$}
    \label{Figure7_2}
\end{subfigure}

\caption{Ablation study on CelebA with different $\sigma$.}
\label{fig7}
\end{figure}

\begin{figure}[ht]
\centering
\begin{subfigure}[b]{0.45\linewidth}
    \includegraphics[width=\linewidth]{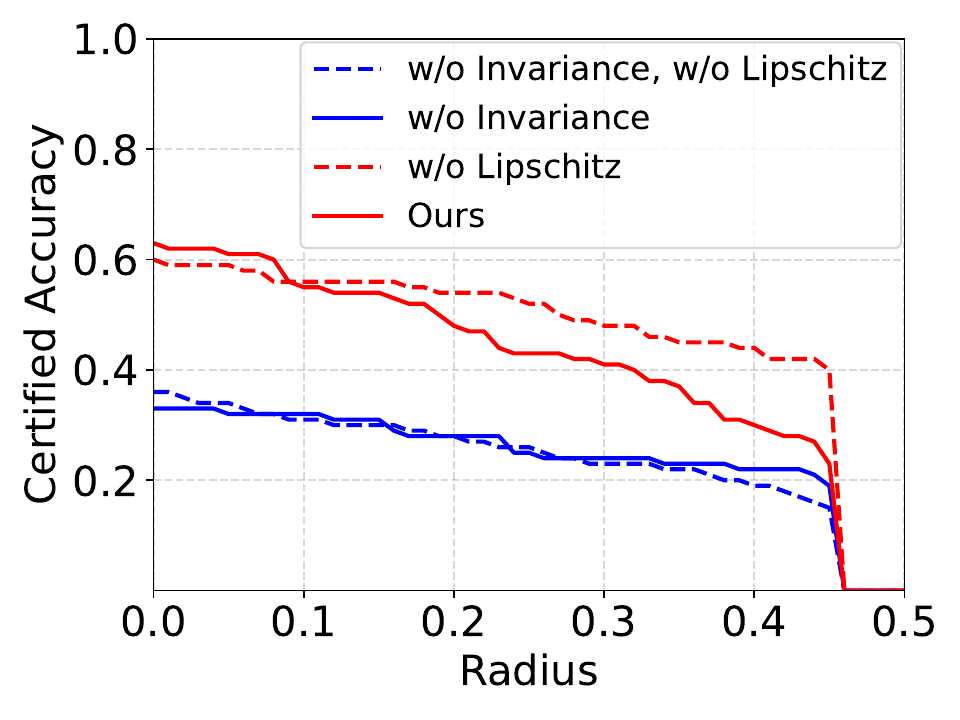}
    \caption{$\sigma=0.12$}
    \label{Figure8_1}
\end{subfigure}
\hfill 
\begin{subfigure}[b]{0.45\linewidth}
    \includegraphics[width=\linewidth]{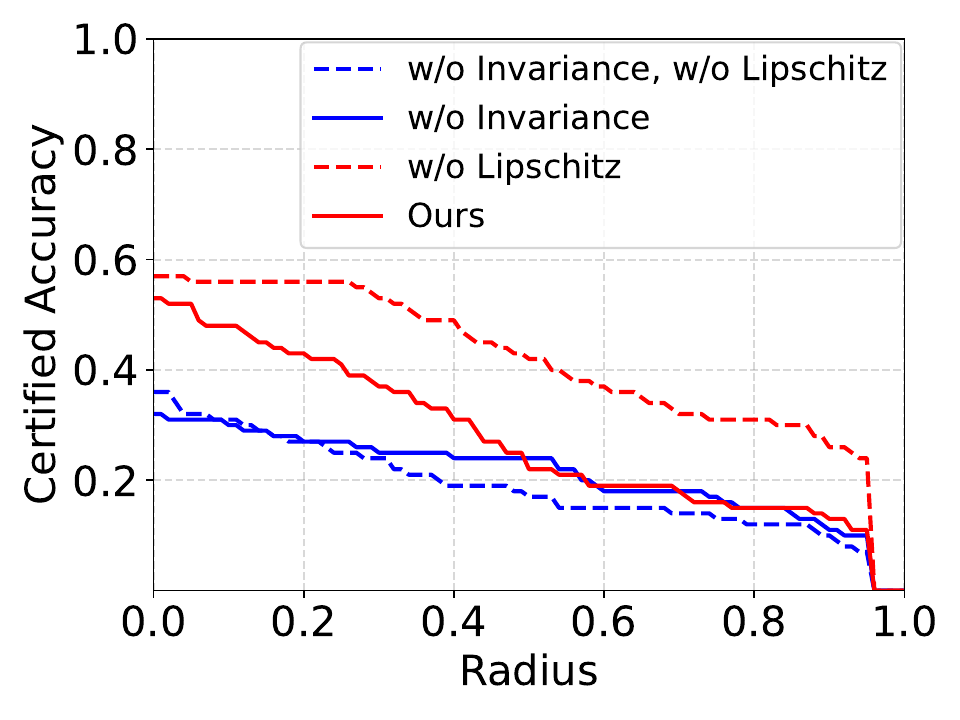}
    \caption{$\sigma=0.25$}
    \label{Figure8_2}
\end{subfigure}

\caption{Ablation study on DomainNet with different $\sigma$.}
\label{fig8}
\end{figure}

\subsubsection{Ablation Study}
\looseness=-1
We conduct the ablation study on CelebA and DomainNet. The results are shown in Figure \ref{fig7} and \ref{fig8}. The results from both datasets demonstrate that our method outperforms the variant without the invariance penalty. For our model on CelebA, the certified accuracy and ACR with the Lipschitz constraint are superior to those without this constraint. Even though on DomainNet this observation is different, it is still reasonable because the Lipschitz constraint limits the learning capability of the neural network, and this limitation is further amplified when dealing with more complex real-world datasets. 
However, without this constraint, there is no guarantee of cross-domain robustness. Even if this variant shows improved certified accuracy, its superiority cannot be proved.

\begin{figure}[ht]
\centering
\begin{subfigure}[b]{0.45\linewidth}
    \includegraphics[width=\linewidth]{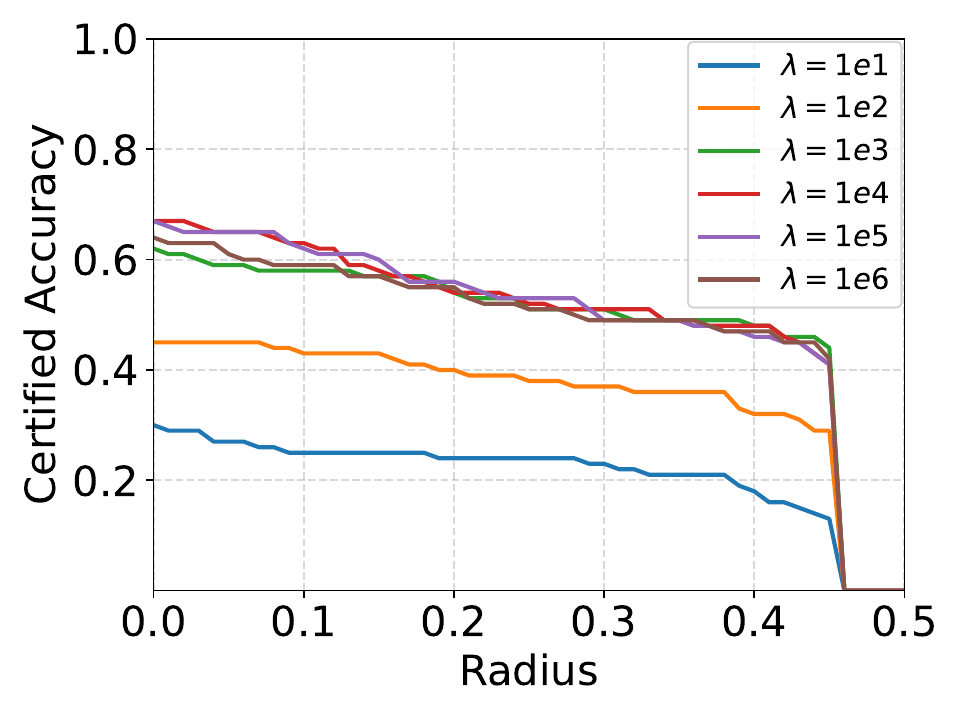}
    \caption{Vary $\lambda$}
    \label{Figure9_1}
\end{subfigure}
\hfill 
\begin{subfigure}[b]{0.45\linewidth}
    \includegraphics[width=\linewidth]{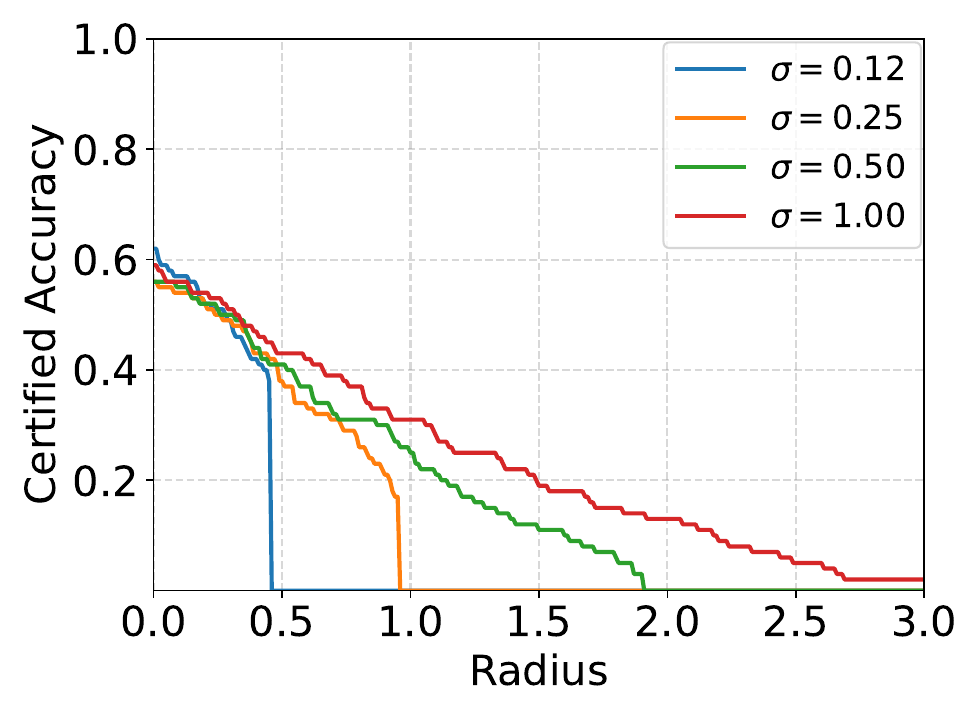}
    \caption{Vary $\sigma$}
    \label{Figure9_2}
\end{subfigure}

\caption{Parameter Study on CelebA.}
\label{fig9}
\end{figure}

\begin{figure}[ht]
\centering
\begin{subfigure}[b]{0.45\linewidth}
    \includegraphics[width=\linewidth]{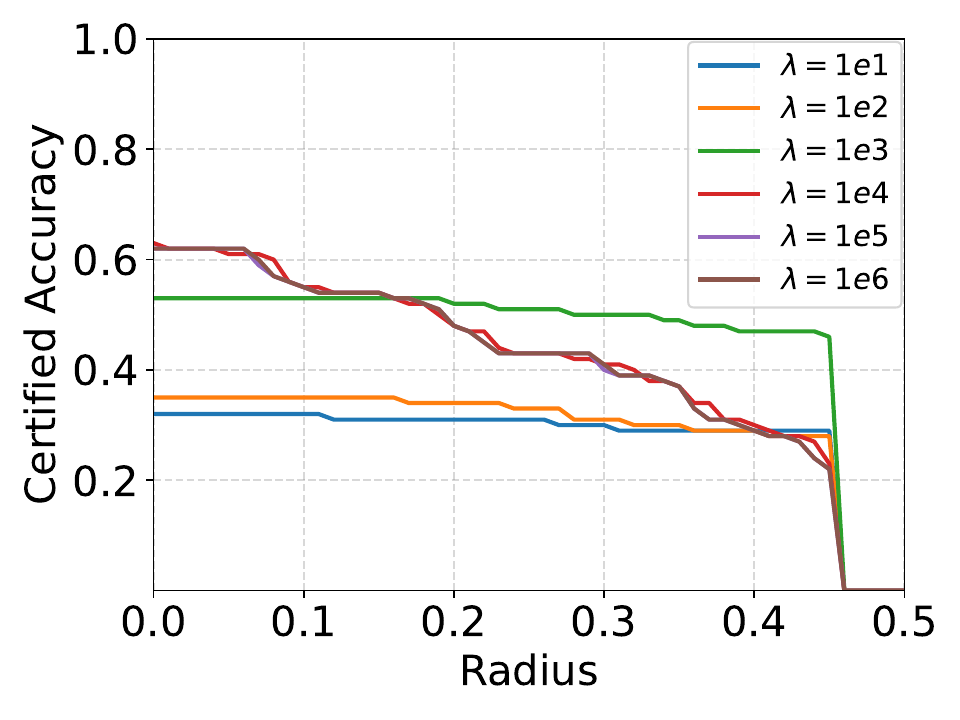}
    \caption{Vary $\lambda$}
    \label{Figure10_1}
\end{subfigure}
\hfill 
\begin{subfigure}[b]{0.45\linewidth}
    \includegraphics[width=\linewidth]{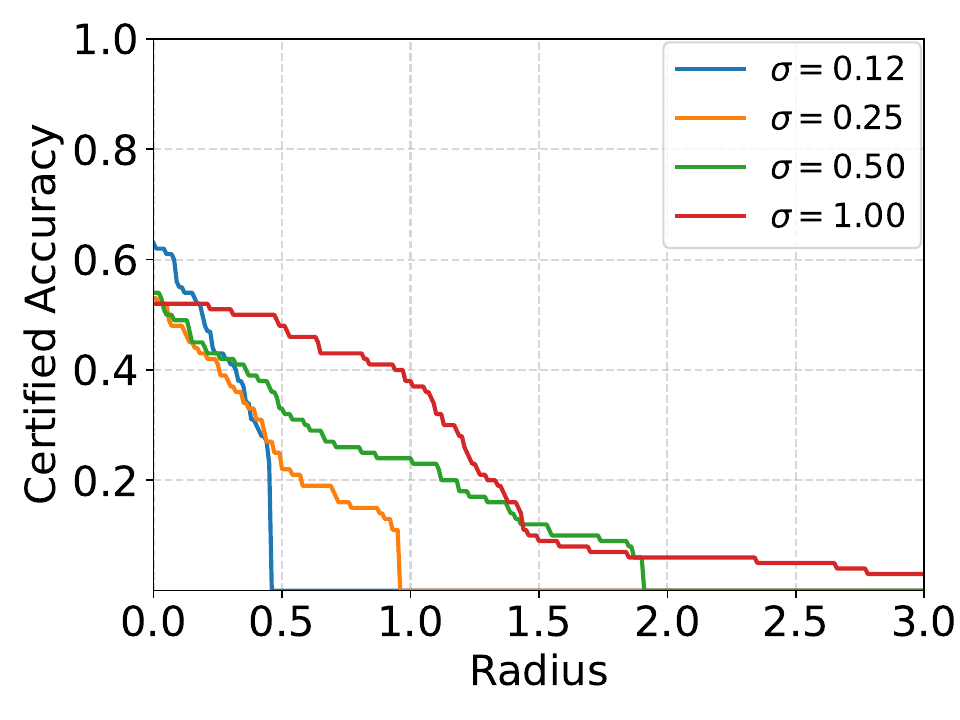}
    \caption{Vary $\sigma$}
    \label{Figure10_2}
\end{subfigure}

\caption{Parameter Study on DomainNet.}
\label{fig10}
\end{figure}
\subsubsection{Parameter Study} We conduct the parameter study on CelebA and DomainNet. Consistent with the main text, we vary $\lambda$ and $\sigma$. The results are shown in Figure \ref{fig9} and Figure \ref{fig10}. For CelebA, the results are similar to those for CMNIST. The only difference is that when $\lambda$ exceeds $10^3$, the model essentially reaches the bottleneck of invariant causal learning. In Figure \ref{fig10}(a), similar to previous analyses, when the dataset is more complex, invariant learning focuses more on generalizability and robustness under small-scale attacks. For Figure \ref{fig10}(b), we still arrive at the following conclusion: higher noise levels can lead to larger certified radii, but as the radius continues to increase, the certified accuracy correspondingly decreases.

\end{document}